\documentclass{article}

\usepackage{microtype}
\usepackage{graphicx}
\usepackage{subfigure}
\usepackage{booktabs} 
\usepackage{hyperref}

\usepackage[accepted]{icml2025}

\usepackage{amsmath}
\usepackage{amssymb}
\usepackage{mathtools}
\usepackage{amsthm}

\usepackage[capitalize,noabbrev]{cleveref}

\theoremstyle{plain}

\theoremstyle{definition}

\theoremstyle{remark}

\usepackage[textsize=tiny]{todonotes}

\icmltitlerunning{Challenges in Understanding Modality Conflict in Vision-Language Models}

\begin{document}

\twocolumn[
\icmltitle{Challenges in Understanding Modality Conflict in Vision-Language Models}



\icmlsetsymbol{equal}{*}

\begin{icmlauthorlist}
\icmlauthor{Trang Nguyen}{equal,umass}
\icmlauthor{Jackson Michaels}{equal,umass}
\icmlauthor{Madalina Fiterau}{umass}
\icmlauthor{David Jensen}{umass}

\end{icmlauthorlist}

\icmlaffiliation{umass}{Manning College of Information \& Computer Sciences, University of Massachusetts Amherst, Amherst, U.S.}
\icmlcorrespondingauthor{Trang Nguyen}{tramnguyen@umass.edu}
\icmlkeywords{Machine Learning, ICML}

\vskip 0.3in
]



\printAffiliationsAndNotice{\icmlEqualContribution} 

\begin{abstract}
This paper highlights the challenge of decomposing conflict detection from conflict resolution in Vision-Language Models (VLMs) and presents potential approaches, including using a supervised metric via linear probes and group-based attention pattern analysis. We conduct a mechanistic investigation of \texttt{LLaVA-OV-7B}, a state-of-the-art VLM that exhibits diverse resolution behaviors when faced with conflicting multimodal inputs. Our results show that a linearly decodable conflict signal emerges in the model’s intermediate layers and that attention patterns associated with conflict detection and resolution diverge at different stages of the network. These findings support the hypothesis that detection and resolution are functionally distinct mechanisms. We discuss how such decomposition enables more actionable interpretability and targeted interventions for improving model robustness in challenging multimodal settings.
\end{abstract}

\section{Introduction}
Recent research has identified a consistent trend in Vision-Language Models (VLMs). When faced with conflicting information between modalities, models often exhibit a bias toward textual input 
\cite{zhangDebiasingMultimodalLarge2024, dengSeeingBelievingMitigating2024}.
The research question that underlies much of this work is: ``\textit{Under what conditions do models tend to favor textual over image information?}''

\textbf{Related Work.}
For example, \citet{dengWordsVisionVisionLanguage2025} finds that token ordering affects modality preference and \citet{dengSeeingBelievingMitigating2024} observes that models are more likely to trust parametric text knowledge when the text is longer or more detailed. \citet{zhangDebiasingMultimodalLarge2024} goes further to suggest that models detect modality conflicts internally, but that the decoding process masks this awareness. However, they did not investigate further the distinction between conflict detection and resolution.

\textbf{Gaps and Opportunities.}
Despite the advances in diagnosing modality preference, none of these studies explore the internal mechanisms that mediate conflict detection and resolution in VLMs. Specifically, there is little understanding of how and where in the network the signals that indicate conflict are processed or resolved. Are there two separate internal processes of detecting and resolving conflicts? If separate processes exist, can they be disentangled? 

Answering this question is pivotal for \textit{actionable} mechanistic interpretability. While the promise of the field is to map understanding of model internals to concrete engineering and practical benefits to AI safety \cite{open_mech_interp}, a persistent criticism has been the difficulty of creating actionable tools from interpretability research especially in Natural Language Processing (NLP) \cite{insight_to_act}. If these mechanisms can be decomposed, it would open the door for targeted interventions and focused monitoring tools to enhance safety and reliability in practical applications. 

\textbf{Our Approach.} Our research shifts the focus from behavioral observations to mechanistic explanations. We ask: ``\textit{Under what circumstances can we decompose and localize conflict detection and resolution mechanisms in VLMs?}'' 

A major challenge in decomposing conflict detection and conflict resolution mechanism is that the commonly used causal intervention effect metric (i.e., an observed outcome like model's output or output probability) cannot distinguish between the two mechanism. This necessitates a metric to measure the intervention effect that can confidently tell us if the model decreases or increases awareness of the conflict internally. However, this is challenging, as we cannot directly observe how the model represents its awareness of the conflict.

This paper explores two approaches to addressing this challenge: (1) using linear probes as a potential metric to measure internal conflict signals; and (2) employing observational comparative methods, such as attention pattern analysis, to identify distinct and shared components of the conflict detection and resolution mechanisms. For each approach, we discuss its practical applications and limitations, focusing not only on mechanistic insights but also on how model users can leverage these insights in practice.

\textbf{Scope.} Within the scope of this study, we only focus on \textit{task-relevant direct modality conflict}. A \textit{task-relevant direct modality conflict} occurs when given a task (i.e., a question), the text provides explicit information for an answer that contradicts what is shown in the image. Future studies will expand to a wider range of conflict, including: task-irrelevant and indirect conflict. 
\vspace{-10pt}

\section{Experiment Setup}
\subsection{Dataset Construction}
\begin{figure}
    \centering
    \includegraphics[width=0.9\linewidth]{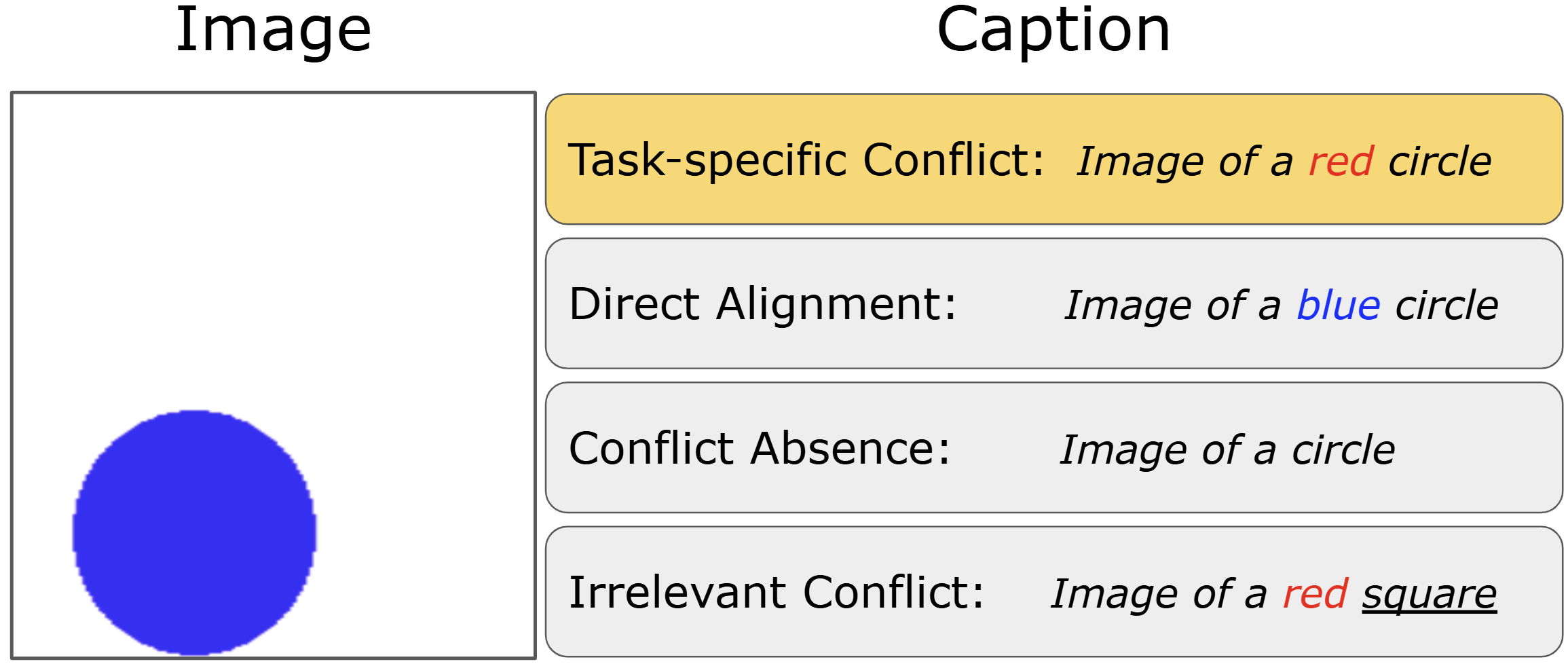}
    \caption{Example synthetic conflict question from generated dataset. Highlighted caption shows base conflict case vs. caption options depending on desired conflict types.} 
    \label{fig:sample_example}
\end{figure}

To avoid confounders present in existing VQA datasets (e.g., vision model capacity, parametric knowledge, and question difficulty) when analyzing modality conflict, we designed a minimal synthetic dataset. 
Each sample pairs a small image (256 $\times$ 256) of a primary-colored shape (e.g., a blue circle) with a caption asserting a conflicting color for that shape (e.g., `an image of a red circle'). 
Figure~\ref{fig:sample_example} shows an example image-caption pair including alternative captions. This design presents a direct modality conflict while minimizing the need for complex logic to identify the conflict.
Our dataset consists of 5,600 samples, evenly distributed across five shapes (circle, square, triangle, diamond, and star) and eight colors (black, gray, red, blue, green, yellow, purple, pink), with seven distinct conflicting caption image pairs for each shape color combination. 
Shapes are randomly placed and sized within the image for varied samples. The dataset is publicly available at \url{https://huggingface.co/datasets/anonymous052025/multimodal-modality-conflict-dataset}.
\vspace{-10pt}

\begin{figure}
    \centering
    \includegraphics[width=0.9\linewidth, height=0.16\textheight]{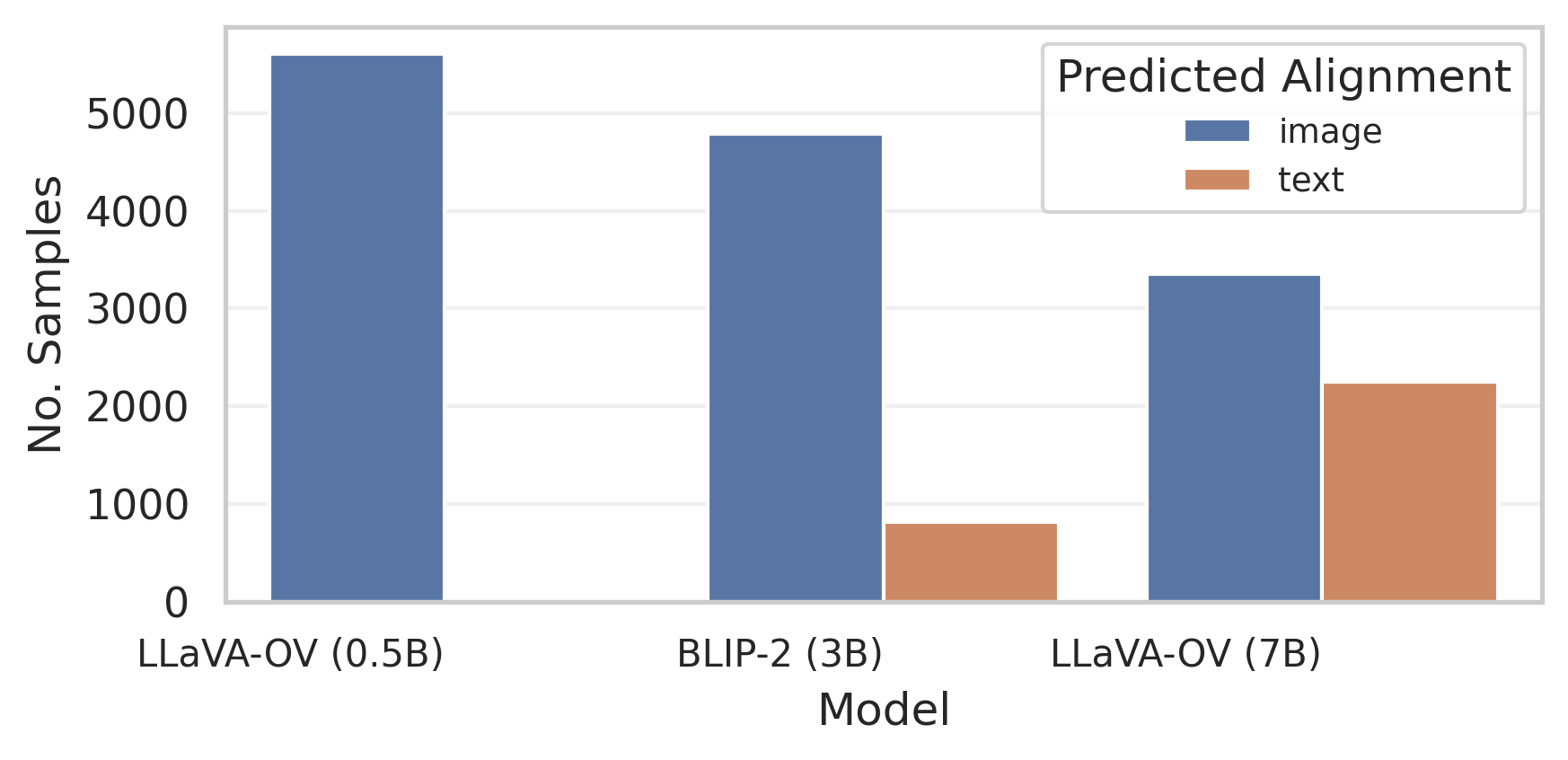}
    \caption{Modality-based Answer Alignment across 3 different VLMs. X-axis is ordered by increasing model size.}
    \label{fig:model_comparison_results}
\end{figure}
\subsection{Models}
We tested three different VLMs of varying sizes, including two models from the LLaVA-OV family \cite{llava_one} and BLIP-2 \cite{li2023blip2}. Figure~\ref{fig:model_comparison_results} illustrates the significant differences in conflict resolution patterns across these models. Given our focus on understanding how to decompose the mechanisms of conflict detection and resolution, we chose to further investigate \texttt{LLaVA-OV-7B} \footnote{\url{https://huggingface.co/lmms-lab/llava-onevision-qwen2-7b-ov}}, as it demonstrates the most balanced preference between image and text in resolving conflicts.
\vspace{-10pt}

\section{Key Insights}
\subsection{The conflict signal becomes linearly detected at intermediate layers, and nonlinearly correlates with VLM conflict resolution confidence}
\label{sec:linear_probe}

\textbf{Setup.}
We train a probe using lasso logistic regression to classify whether a data instance contains a \textit{task-relevant modality conflict} using layerwise last-token activations of the VLM. Table~\ref{tab:linear_probe_setup} summarizes description and justification for key design choices of our setup.

\begin{table*}[t]
\centering
\caption{Design choices to rule out spurious cues in conflict detection probing.}
\label{tab:linear_probe_setup}
\begin{tabular}{p{0.28\textwidth} p{0.28\textwidth} p{0.40\textwidth}}
\toprule
\textbf{Design Choice} & \textbf{Confounding Factor Ruled Out} & \textbf{Rationale} \\
\midrule
Disjoint train/test color sets 
& Memorizing color
& Prevents probe from only learning specific colors \\ 
\addlinespace
No-conflict samples with no color and matching color captions
& Checking the existence of color token in caption
& Ensures probe does not merely check existence of color token in caption to determine conflict \\ 
\addlinespace
No-conflict samples with caption color for a different shape 
& Counting number of color tokens 
& Ensures probe do not just classify all samples with multiple different colors as conflict \\ 
\bottomrule
\end{tabular}
\end{table*}

\begin{figure}
    \centering
    \includegraphics[width=\linewidth]{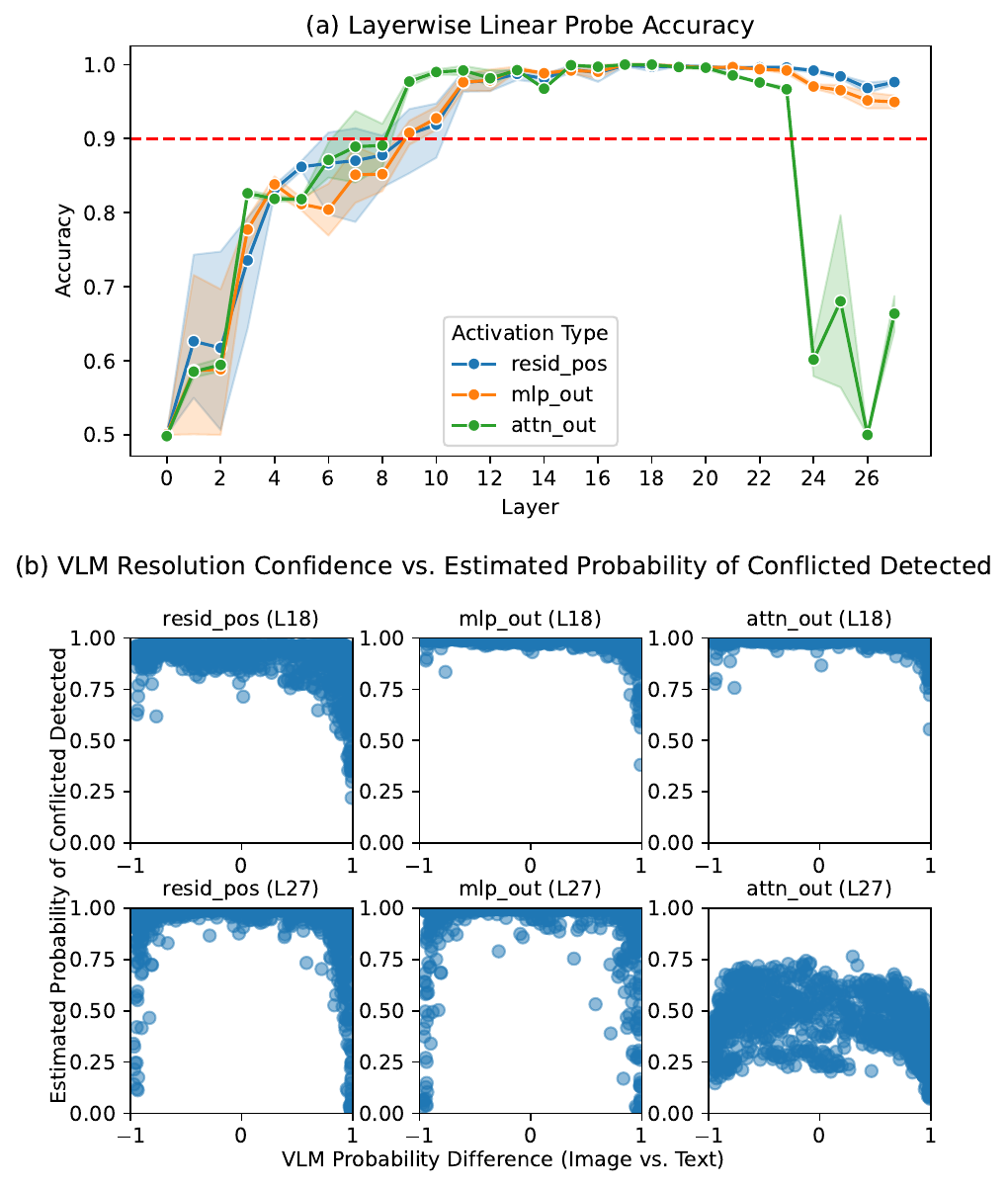}
    \caption{\textbf{Performance of conflict detection linear probe and its relationship to the VLM's conflict resolution confidence.} (a) Linear probe accuracy across layers, showing when conflict signal becomes linearly detected. (b) Correlation between the linear probe's estimated probability of conflict and the VLM's conflict resolution confidence. Resolution confidence is measured as the probability difference between image-based and text-based answers, ranging from $-1.0$ to $1.0$; values near $-1.0$ or $1.0$ indicate strong confidence in one modality, while values near $0$ indicate uncertainty. All results reported here are on the test set.}

    \label{fig:linear_probe}
\end{figure}

\textbf{Results.}
As shown in Figure~\ref{fig:linear_probe}(a), starting at layer 10, linear probes trained on all three activation types—attention output, MLP output, and residual stream—achieve high accuracy in distinguishing between conflict and no-conflict samples. This suggests that a linearly decodable conflict signal emerges in the model's intermediate hidden states, especially in the middle layers. Notably, probe accuracy for attention output declines in the final few layers, suggesting that later attention layers no longer maintain a linearly accessible conflict representation. This may reflect a functional shift, where late-stage attention is repurposed for conflict resolution-specific processes rather than conflict detection. Consistent with this interpretation, our analysis of attention patterns in Section~\ref{sec:attn_pattern} shows that divergence in attention patterns related to conflict detection predominantly emerges before those related to conflict resolution. 

Assuming the presence of a conflict detection subspace, we next examine its relationship to the model's conflict resolution behavior—specifically, how the model chooses between image-based and text-based answers when presented with conflicting inputs. We choose to quantify the model's resolution confidence using probability difference because probability captures the model's relative internal preference between the two options \cite{wiegreffeanswer}, and is appropriate in our setting since the expected answers are mutually exclusive, avoiding issues related to semantic overlap between answer choices \cite{hu2023uncertainty}. The strength of the internal conflict detection signal is measured by the estimated probability of conflict presence, as predicted by a fitted linear probe.
Figure~\ref{fig:linear_probe}(b) shows a nonlinear correlation and a heteroscedastic relationship between the VLM's conflict resolution confidence and the conflict strength. When the model is uncertain between the two answers (confidence closer to $0$), the internal conflict detection signal tends to be strong and consistent. In contrast, when the model is highly confident in its resolution decision (confidence near $-1$ or $1$), the mean conflict detection strength declines, and its variance increases. 

Furthermore, we observe that when the conflict detection signal is strong, the model exhibits a wide range of resolution behaviors—from uncertainty to strong preference for either modality. This variance implies that resolution decisions are not tightly determined by the detection signal alone. While the model may detect conflict internally, it resolves conflict in diverse ways. These findings suggest that conflict detection and resolution mechanisms can be decoupled.

\textbf{Takeaway and Limitation.}
Our results suggest that task-relevant modality conflict is explicitly encoded in the VLM's intermediate activations. This evidence supports the hypothesis that conflict detection is a distinct and localizable mechanism within the network. Practically, this insight enables efficient conflict monitoring in production systems: by accessing the model's intermediate activations, conflict signals can be detected in a single forward pass, without requiring complex prompting or behavioral probing.

However, despite its effectiveness, linear probe only detects this signal within a linear subspace, which may overlook more subtle representations of conflict within the model. Additionally, the existence of this signal does not directly inform us about how, or whether, it influences the model’s resolution decisions. To fully understand how the model operationalizes detected conflict in its decision-making processes, further causal analyses and targeted interventions using this as an intervention metric are necessary.

\subsection{Group Based Attention Head Analysis Reveals Distinct Patterns for Detection vs. Resolution} 
\label{sec:attn_pattern}
\textbf{Setup.}
Motivated by previous works, such as \citet{induct_head}, we analyzed the behavior of the model's attention heads to investigate how the model internally differentiates between detecting and resolving a conflict. We hypothesized that distinct attention heads might specialize in these two sub-processes. For each sample, we extracted attention pattern from the model’s final output token to the text color token and all image tokens. 

To distinguish between conflict detection and resolution mechanisms, we compare the change in attention patterns within two groups: 
(1) \emph{Conflict vs. No Conflict (Detection Mechanism):} We compared average attention patterns from samples containing conflicting image-text information against patterns from non-conflicting samples; 
(2) \emph{Text-aligned vs. Image-aligned (Resolution Mechanism):} Focusing only on conflicting samples, we compared average attention patterns from instances where the model's final output aligned with the image against those aligned with the text. We then calculated the difference between these averaged attention patterns for each comparison pair (i.e., conflict vs. no-conflict, and image-aligned vs. text-aligned). This difference analysis aimed to identify attention heads showing significant changes in behavior corresponding to either detection or resolution processes.

\textbf{Results.}
\begin{figure}
    \centering
    \includegraphics[width=1\linewidth]{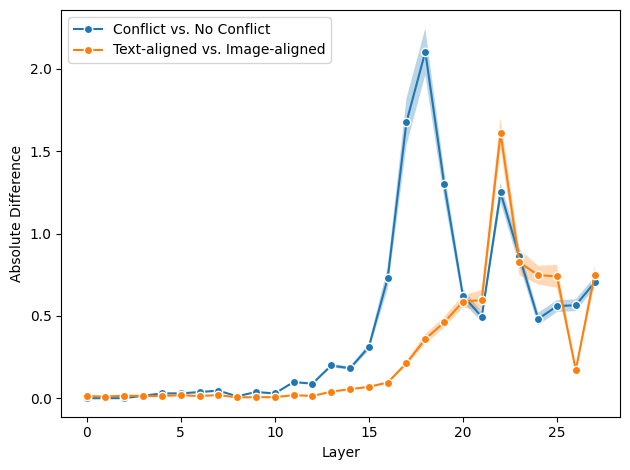}
\caption{Layer-wise absolute difference in attention weights from final token to text color token in each group. Attention weights are summed across heads in each layer. Shaded regions show standard deviation. Detection behavior peaks earlier than resolution, both primarily post-layer 15.}
    \label{fig:attn_act}
\end{figure}
Our analysis revealed distinct differences in attention allocation, particularly on a small subset of attention heads primarily located after layer 15, as shown in Figure~\ref{fig:attn_act}. This divergence suggests that these specific heads play a key role in the conflict resolution or detection of the model mediating the final selection between conflicting modalities before output. These results indicate two mostly non-overlapping clusters: one that differentiates conflict vs.\ no-conflict samples, and one that differentiates image-aligned vs.\ text-aligned resolutions. Most importantly, this analysis also show attention changes in the detection mechanism before the resolution mechanism.

\textbf{Takeaway and Limitation.}
Group-based attention analysis shows that these two components can likely be disentangled and localized to a small subset of attention heads. 
While we can identify mechanism-specific components using correlational evidence, this method does not provide causal evidence on the minimal or sufficient effect of these components on each mechanism. 
Any causal method like patching will require a mechanism-specific intervention metric.
Future work will use a linear probe as a potential metric for the conflict detection mechanism and model output as a metric for the alignment mechanism.

\subsection{Conclusion}
This study presents our first attempt to decompose the mechanisms of conflict detection and resolution by conducting a mechanistic analysis of a state-of-the-art Vision-Language Model, \texttt{LLaVA-OV-7B}, which exhibits diverse resolution behavior between image- and text-based modalities. Our results suggest the existence of a linearly decodable subspace in the model’s intermediate layers that encodes task-specific conflict signals, providing a concrete handle for monitoring internal conflict detection. Additionally, we observe distinct attention patterns associated with conflict detection and resolution, with divergence in detection emerging earlier in the network than resolution behavior—supporting the hypothesis that these are functionally and temporally separate processes. Decomposing these mechanisms, and subsequently tailoring interventions (e.g., prompting or patching) to target each one independently, becomes feasible by using the linear probe–based conflict detection estimate as a proxy metric for internal detection. This enables researchers and practitioners to disentangle whether an intervention affects conflict detection, resolution, or both—offering a more actionable path toward interpretability in real-world applications such as model monitoring and robustness under modality conflict.

However, while the linear probe’s estimated conflict probability provides a promising supervised metric for accessing the internal conflict signal, identifying robust and interpretable metrics in the unsupervised setting remains an open challenge for future work.

\bibliography{manual_references}
\bibliographystyle{icml2025}
\end{document}